\documentclass{article}

\usepackage{microtype}
\usepackage{graphicx}
\usepackage{subcaption}
\usepackage{booktabs} % for professional tables

\usepackage{hyperref}

\usepackage[accepted]{icml2025}   %  [accepted]

\usepackage{amsmath}
\usepackage{amssymb}
\usepackage{mathtools}
\usepackage{amsthm}

\usepackage[capitalize,noabbrev]{cleveref}

\theoremstyle{plain}

\theoremstyle{definition}

\theoremstyle{remark}

\usepackage[textsize=tiny]{todonotes}

\icmltitlerunning{Element-wise Attention Is All You Need}

\begin{document}

\twocolumn[
\icmltitle{Element-wise Attention Is All You Need}



\icmlsetsymbol{equal}{*}

\begin{icmlauthorlist}
\icmlauthor{Guoxin Feng}{yyy}
%\icmlauthor{Firstname2 Lastname2}{equal,yyy,comp}
%\icmlauthor{Firstname3 Lastname3}{comp}
%\icmlauthor{Firstname4 Lastname4}{sch}
%\icmlauthor{Firstname5 Lastname5}{yyy}
%\icmlauthor{Firstname6 Lastname6}{sch,yyy,comp}
%\icmlauthor{Firstname7 Lastname7}{comp}
%\icmlauthor{Firstname8 Lastname8}{sch}
%\icmlauthor{Firstname8 Lastname8}{yyy,comp}
\end{icmlauthorlist}

\icmlaffiliation{yyy}{College of Information Science and Engineering, Northeastern University, China}
%\icmlaffiliation{comp}{Company Name, Location, Country}
%\icmlaffiliation{sch}{School of ZZZ, Institute of WWW, Location, Country}

\icmlcorrespondingauthor{Guoxin Feng}{2172064@stu.neu.edu.cn}
%\icmlcorrespondingauthor{Firstname2 %Lastname2}{first2.last2@www.uk}

\icmlkeywords{}

\vskip 0.3in
]



\printAffiliationsAndNotice{}

\begin{abstract}
The self-attention (SA) mechanism has demonstrated superior performance across various domains, yet it suffers from substantial complexity during both training and inference. The next-generation architecture, aiming at retaining the competitive performance of SA while achieving low-cost inference and efficient long-sequence training, primarily focuses on three approaches: linear attention, linear RNNs, and state space models. Although these approaches achieve reduced complexity than SA, they all have built-in performance degradation factors, such as diminished “spikiness” and compression of historical information. In contrast to these approaches, we propose a novel element-wise attention mechanism, which uses the element-wise squared Euclidean distance, instead of the dot product operation, to compute similarity and approximates the quadratic complexity term $exp(q_{ic}k_{jc})$ with a Taylor polynomial. This design achieves remarkable efficiency: during training, the element-wise attention has a complexity of $\mathcal{O}(tLD)$, making long-sequence training both computationally and memory efficient, where $L$ is the sequence length, $D$ is the feature dimension, and $t$ is the highest order of the polynomial; during inference, it can be reformulated as recurrent neural networks, achieving a inference complexity of $\mathcal{O}(tD)$. Furthermore, the element-wise attention circumvents the performance degradation factors present in these approaches and achieves performance comparable to SA in both causal and non-causal forms. 
\end{abstract}

\section{Introduction}
\label{submission}
Traditional deep learning methods have significant limitations: Convolutional Neural Networks (CNNs) struggle to capture long-range dependencies, while traditional Recurrent Neural Networks (RNNs) \cite{lstm,gru} do not support parallel training and are prone to exploding or vanishing gradients. In contrast, the self-attention (SA) \cite{SA} mechanism overcomes these limitations and has demonstrated superior performance across various domains \cite{Transformer-XL,vit,gpt3}, including natural language processing (NLP), computer vision (CV), and time-series analysis.

Despite its superior performance, SA incurs significant complexity in both training and inference. During inference, the KV-caching \cite{KV-caching} technique stores historical key and value matrices, leading to an inference complexity of $\mathcal{O}(LD)$, where $L$ is the sequence length and $D$ is the feature dimension. Consequently, the inference cost grows with the sequence length. During training, SA has computational and memory complexities of $\mathcal{O}(L^2)$, which makes training on long sequences both time-consuming and memory-intensive. 

Significant efforts have been devoted to developing the next-generation architecture, aiming at simultaneously achieving low-cost inference, efficient long-sequence training, and SA-comparable performance, i.e., breaking the “impossible triangle”. Current research can be classified into three main categories. First, linear attention (LA) \cite{LA,retnet,DeltaNet,gla} employs kernels $\varphi(q_i) \cdot \varphi(k_j^T)$ to approximate the $exp(q_ik_j^T)$ in SA, achieving linear training complexity and constant inference complexity with respect to the sequence length. However, LA's performance often lags behind SA, primarily due to its lack of "spikiness" \cite{hedgehog}. Spikiness refers to the attention mechanisms assigning major weights to a few critical tokens while assigning little weights to others, leading to sharp attention weights distributions, which greatly contributes to the effectiveness of attention mechanisms. SA attains spikiness by using the exponential function to amplify larger dot products and suppress smaller ones. In contrast, LA loses this spikiness by eliminating the exponential function. Secondly, linear RNNs \cite{qrnn,lru,rwkv} eliminate the non-linear functions in traditional RNNs, enabling efficient parallel training while retaining the low-cost inference characteristic in traditional RNNs. However, linear RNNs generally exhibit inferior performance compared to attention mechanisms. One reason is that eliminating non-linear functions decreases the models' representational capacity. Another reason is that, unlike attention mechanisms allowing tokens to interact directly, RNNs process data sequentially and compress historical information into a single hidden vector. This process gradually diminishes prior information, limiting the models' ability to capture long-range dependencies. The third category of research focuses on simulating state space models (SSMs) \cite{s4,h3,mamba,Hyena,rft}. These models exhibit performance degradation factors similar to those in RNNs, and require complex implementations to maintain efficiency during training and inference. None of the three categories of work has yet succeeded in breaking the “impossible triangle”.

Research \cite{based} has demonstrated that the Taylor polynomial can preserve the spikiness introduced by the exponential function ($exp(q_i k_j^\top) = 1 + q_i k_j^\top + \frac{(q_i k_j^\top)^2}{2} + \cdots$), leading to the training complexity of $\mathcal{O}(LD^t)$ and the inference complexity of $\mathcal{O}(D^t)$, where $t$ denotes the highest order of the polynomial. Notably, the exponential complexity with respect to the feature dimension $D$ arises from the dot-product operation, which is the standard operation for computing similarity in traditional attention mechanisms. We propose that, by replacing the dot-product operation with element-wise operation, the incredible efficiency (a training complexity of $\mathcal{O}(tLD)$ and an inference complexity of $\mathcal{O}(tD)$) can be achieved. And the proposed element-wise attention avoids the aforementioned performance degradation factors. Specifically, we use the element-wise squared Euclidean distance to measure the similarity between query and key elements at an arbitrary channel $c$. Then, the Softmax function converts the similarity scores into attention weights. The quadratic complexity term $exp(q_{ic}k_{jc})$ introduced by the Softmax function is approximated using a Taylor polynomial. 

We conducted extensive experiments to compare the element-wise attention with SA, and the results demonstrate that the element-wise attention simultaneously achieves low-cost inference, efficient long-sequence training, and SA-comparable performance. In performance comparisons, the element-wise attention achieves results comparable to self-attention in both causal and non-causal forms on time series datasets, particularly when high-order Taylor polynomials are applied. During training, the element-wise attention demonstrates higher training efficiency than SA in terms of memory usage and throughput. And this advantage becomes more pronounced as the sequence length increases. During inference, the element-wise attention can be reformulated as recurrent neural networks, resulting in inference costs that remain constant with respect to the sequence length and are minimally impacted by variations in batch size. In contrast, SA relies on KV-caching, leading to inference costs that scale with sequence length and are significantly affected by changes in batch size.

\section{Notations}

Let \( X \in \mathbb{R}^{L \times D} \) denote a sequence of \( L \) token vectors, each comprising \( D \) channels. We introduce the following notation: \( X_{i:} \in \mathbb{R}^D \) or \( X_i \in \mathbb{R}^D \) denotes the \( i \)-th token vector; \( X_{:c} \in \mathbb{R}^L \) denotes the elements in the \( c \)-th channel across all tokens; \( X_{ic} \in \mathbb{R} \) denotes the element at the \( i \)-th token vector and \( c \)-th channel.

In Self-Attention (SA), dot-product operations are frequently used. Specifically, for \( q_i = (q_{i1}, \dots, q_{iD}) \in \mathbb{R}^D \) and \( k_j = (k_{j1}, \dots, k_{jD}) \in \mathbb{R}^D \), their dot product is calculated as $q_ik_j^T=q_{i1}k_{j1}+\dots+q_{iD}k_{jD} \in \mathbb{R}$. In Element-wise Attention (EA), element-wise operations are used. Specifically, the element-wise squaring of \( q_i \) is $q_i^2 = \left( q_{i1}^2, \dots, q_{iD}^2 \right) \in \mathbb{R}^D$; the element-wise multiplication between \( q_i \) and \( k_j \) is $q_ik_j = \left( q_{i1} k_{j1}, \dots, q_{iD} k_{jD} \right) \in \mathbb{R}^D$; and the element-wise division is given by $\frac{q_i}{k_j} = \left( \frac{q_{i1}}{k_{j1}}, \dots, \frac{q_{iD}}{k_{jD}} \right) \in \mathbb{R}^D$.

\section{Element-wise Attention}

In section \ref{sec:EA}, we formalize the full version of Element-wise Attention (EA), which can serve as a replacement of SA in Transformer architectures. Subsequently, in section \ref{sec:EA-series}, the EA-series, with only linear training complexity, is derived by approximating the quadratic complexity term in EA using a Taylor polynomial. In section \ref{sec:Causal EA-series}, we formalize the causal EA-series, which can be reformulated as recurrent neural networks (RNNs) to enable efficient inference. Finally, in section \ref{sec:Relation to and Differences from Previous Methods}, EA and EA-series are compared with various methods.

\begin{figure*}[ht]
	\vskip 0.2in
	\begin{center}
		\centerline{\includegraphics[width=\textwidth]{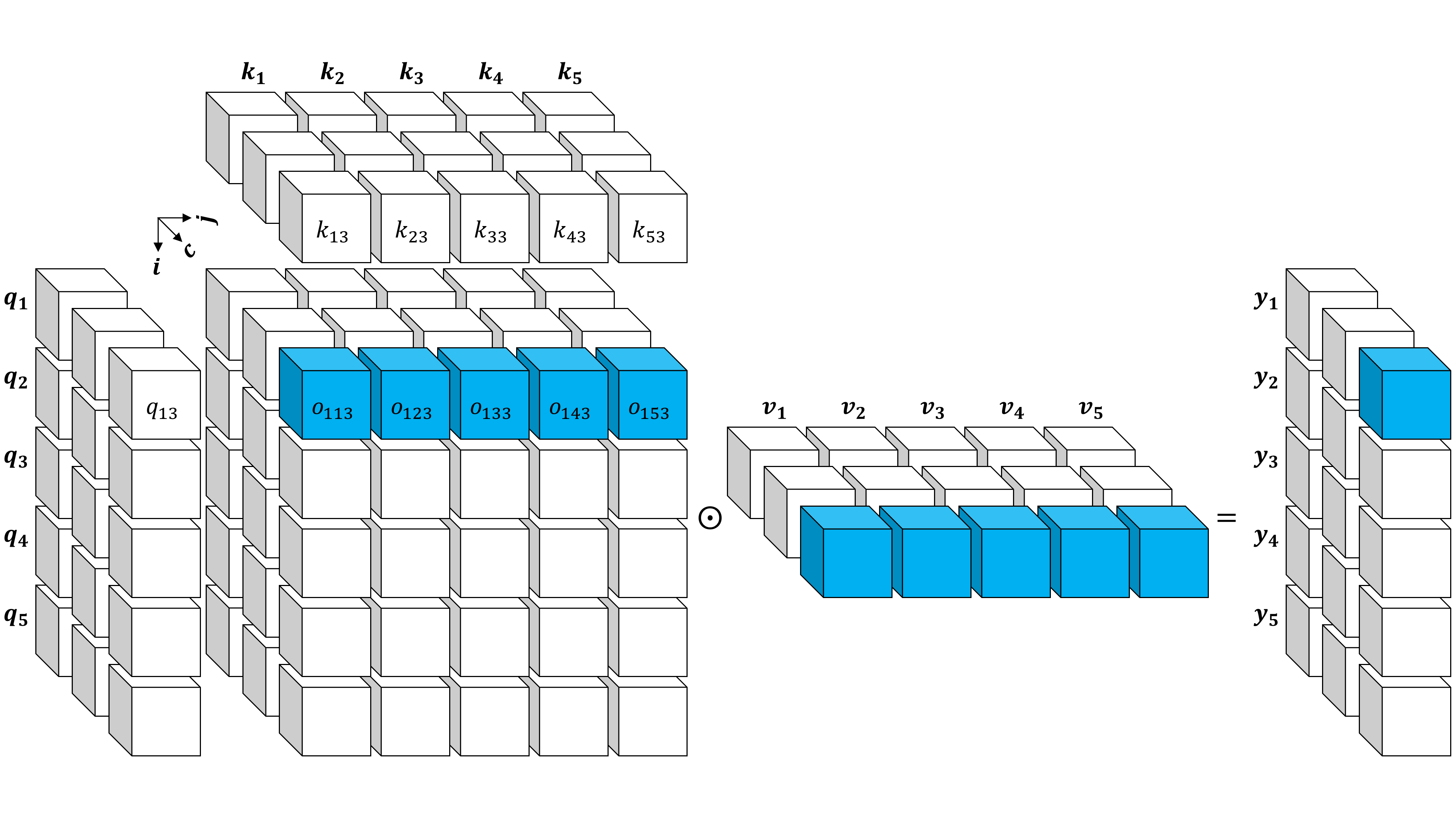}}
		\caption{Illustration of EA’s computation process. Specifically, we obtain the similarity scores $o_{ijc}\in\mathbb{R}$ by computing the squared Euclidean distances between the query element $q_{ic}\in\mathbb{R}$ and the key element $k_{jc}\in\mathbb{R}$. Subsequently, the Softmax function converts $o_{i:c}\in\mathbb{R}^L$ into weights, which are assigned to $v_{:c}\in\mathbb{R}^L$ to produce $y_{ic}\in\mathbb{R}$.}
		\label{fig:Computation Process of EA}
	\end{center}
	\vskip -0.2in
\end{figure*}

\subsection{EA} \label{sec:EA}

In this section, we introduce the full version of EA. Figure \ref{fig:Computation Process of EA} illustrates the computation process of EA.

Specifically, the input sequence \( X \in \mathbb{R}^{L \times D} \) is first projected onto the corresponding representations: \textit{queries} ($q$), \textit{keys} ($k$), \textit{values} ($v$) $\in \mathbb{R}^{L \times D}$. Then, a feature tensor is constructed using $q$ and $k$. The element $o_{ijc} \in \mathbb{R}$ in the feature tensor, which directly quantifies the similarity between $q_{ic} \in \mathbb{R}$ and $k_{jc} \in \mathbb{R}$, is calculated as follows:
\begin{equation}
	o_{ijc}=-(q_{ic}-k_{jc})^2
	\label{similarity in EA}
\end{equation}
Subsequently, a Softmax function normalizes $o_{i:c}\in\mathbb{R}^L$, transforming them into weights that are then assigned to $v_{:c}\in\mathbb{R}^L$ to produce $y_{ic}\in\mathbb{R}$. 

The complete computational formula for EA is presented below, where squaring, multiplication and division operations are all performed element-wise: 
\begin{equation}
	y_i = \frac{\sum_{j=1}^{L} e^{-(q_i - k_j)^2} v_j}{\sum_{j=1}^{L} e^{-(q_i - k_j)^2}}
	\label{the complete computational formula for EA}
\end{equation}
\subsection{EA-series } \label{sec:EA-series}
EA construct a feature tensor to meticulously mobilizes the value elements, but it introduces significant complexity. To reduce the complexity and facilitate efficient training and inference, we approximate the quadratic complexity term in EA using a Taylor polynomial, deriving the EA-series. The derivation process is outlined below. 

\begin{figure}[t]
	\vskip 0.2in
	\begin{center}
		\centerline{\includegraphics[width=\columnwidth]{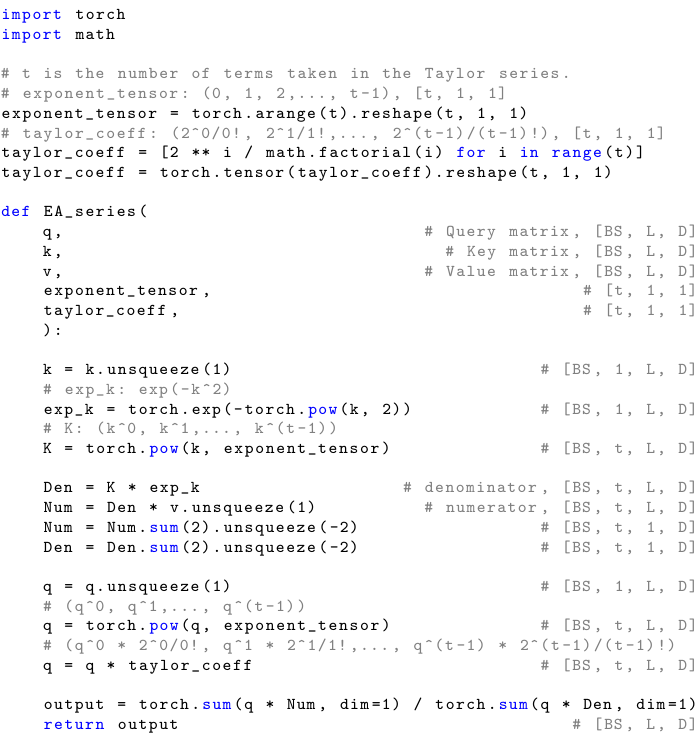}}
		\caption{PyTorch implementation of EA-series.}
		\label{Pytorch_Implementation_of_EA_series}
	\end{center}
	\vskip -0.2in
\end{figure}

Given that the squaring, multiplication and division operations are performed element-wise, equation \ref{the complete computational formula for EA} can be reformulated as follows: 
\begin{align}
y_i &= \frac{\sum_{j=1}^{L} e^{-(q_i - k_j)^2} v_j}{\sum_{j=1}^{L} e^{-(q_i - k_j)^2}} \notag \\
	&= \frac{\sum_{j=1}^{L} e^{-k_j^2} e^{-q_i^2} e^{2q_i k_j} v_j}{\sum_{j=1}^{L} e^{-k_j^2} e^{-q_i^2} e^{2q_i k_j}} \notag \\
	&= \frac{\sum_{j=1}^{L} e^{-k_j^2} e^{2q_i k_j} v_j}{\sum_{j=1}^{L} e^{-k_j^2} e^{2q_i k_j}}
	\label{reformulated computational formula for EA}
\end{align}
We can find that the quadratic complexity arises from computing $e^{2q_ik_j}$. To reduce the complexity, we approximate $e^{2q_ik_j}$ using a Taylor polynomial: 
\begin{equation}
e^{2q_i k_j} = \sum_{n=0}^{+\infty} \frac{(2q_i k_j)^n}{n!} = 1 + 2q_i k_j + \frac{2^2}{2!} q_i^2 k_j^2 + \cdots
	\label{a Taylor polynomial}
\end{equation}
Substituting this polynomial into equation \ref{reformulated computational formula for EA}, we can obtain the EA-series with only linear complexity:
\begin{equation*}
	y_i = \frac{\sum_{j=1}^L e^{-k_j^2} e^{2q_i k_j} v_j}{\sum_{j=1}^L e^{-k_j^2} e^{2q_i k_j}} 
\end{equation*}
\begin{equation*}
	= \frac{\sum_{j=1}^L \left[ e^{-k_j^2} v_j \left( 1 + 2q_i k_j + \frac{2^2}{2!} q_i^2 k_j^2 + \cdots \right) \right]}{\sum_{j=1}^L \left[ e^{-k_j^2} \left( 1 + 2q_i k_j + \frac{2^2}{2!} q_i^2 k_j^2 + \cdots \right) \right]}
\end{equation*}	
\begin{equation*}
	= \frac{\sum_{j=1}^L \left[ e^{-k_j^2} v_j + 2q_i k_j e^{-k_j^2} v_j + \frac{2^2}{2!} q_i^2 k_j^2 e^{-k_j^2} v_j + \cdots \right]}{\sum_{j=1}^L \left[ e^{-k_j^2} + 2q_i k_j e^{-k_j^2} + \frac{2^2}{2!} q_i^2 k_j^2 e^{-k_j^2} + \cdots \right]}
\end{equation*}	
\begin{equation}
	= \frac{\sum_{j=1}^L e^{-k_j^2} v_j + 2q_i \sum_{j=1}^L k_j e^{-k_j^2} v_j + \cdots}{\sum_{j=1}^L e^{-k_j^2} + 2q_i \sum_{j=1}^L k_j e^{-k_j^2} + \cdots}
	\label{non-causal EA-series} 
\end{equation}	
Despite showing the complex form, the EA-series can be efficiently implemented using the broadcast mechanism in PyTorch. The PyTorch implementation is shown in figure \ref{Pytorch_Implementation_of_EA_series}. 

\begin{figure}[t]
	\vskip 0.2in
	\begin{center}
		\centerline{\includegraphics[width=\columnwidth]{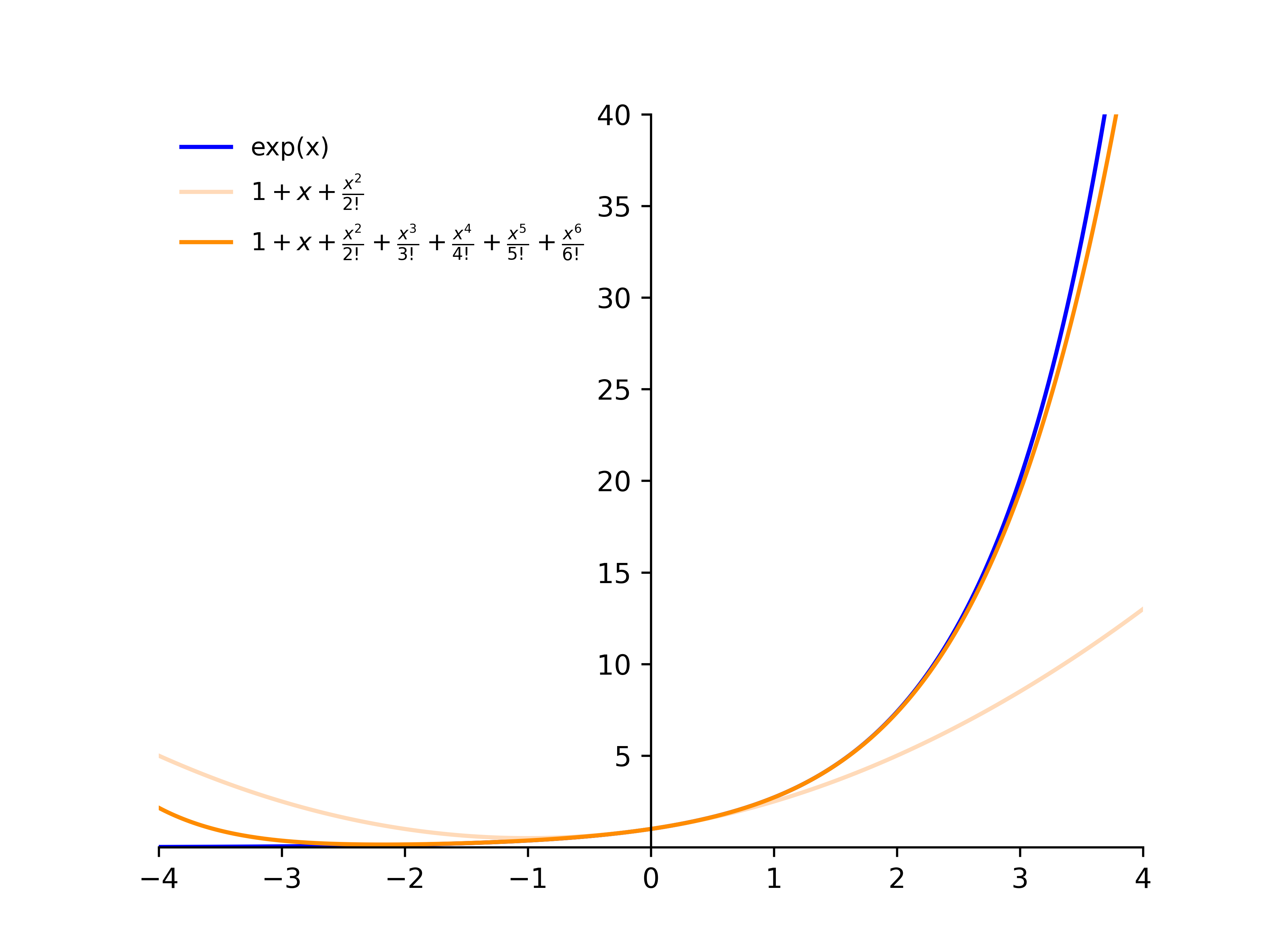}}
		\caption{An illustration of $e^x$ alongside its second- and sixth-order Taylor polynomials.}
		\label{exp_alongside its second- and sixth-order Taylor polynomials}
	\end{center}
	\vskip -0.2in
\end{figure}

\textbf{$\boldsymbol{t}$-th order Taylor polynomial} There are two considerations when selecting a $t$-th order Taylor polynomial. Firstly, the $e^{-{k_j}^2}e^{2q_ik_j}$ in equation \ref{reformulated computational formula for EA}, which serves as similarity scores, must be positive definite. This requires that the polynomial approximation of $e^{2q_ik_j}$ must be positive definite. Research \cite{banerjee2020exploring} has demonstrated that, when $t$ is even, the Taylor polynomial approximation of an exponential function is always positive definite. The second consideration involves the errors between the exponential function and its Taylor polynomial approximations. Figure \ref{exp_alongside its second- and sixth-order Taylor polynomials} illustrates $e^x$ alongside its second- and sixth-order Taylor polynomials. Near the origin, the errors are minimal and decrease as additional terms are incorporated. Errors at points far from the origin are not problematic, because intermediate variables typically remain within a restricted range near the origin due to initialization and normalization techniques. 

\subsection{Causal EA-series } \label{sec:Causal EA-series}

Attention mechanisms have two forms: causal and non-causal. In non-causal attention, each token can attend to all tokens within the sequence. Equation \ref{non-causal EA-series} is the non-causal EA-series. Conversely, causal attention allows each token to attend only to preceding and current tokens, excluding subsequent tokens. Large language models (LLMs) typically use causal attention. The causal EA-series is formally defined in equation \ref{causal EA-series} and can be implemented by replacing the \emph{sum()} in figure \ref{Pytorch_Implementation_of_EA_series} with \emph{cumsum()}.
\begin{equation}
y_i = 
\frac{
	\sum_{j=1}^i e^{-k_j^2} v_j + 2q_i \sum_{j=1}^i k_j e^{-k_j^2} v_j 
	+ \cdots
}{
	\sum_{j=1}^i e^{-k_j^2} + 2q_i \sum_{j=1}^i k_j e^{-k_j^2} 
	+ \cdots
}
\label{causal EA-series}
\end{equation}
The causal EA-series can be trained in parallel with linear complexity. When it comes to inference, the causal EA-series can be rewritten as recurrent neural networks:
\begin{equation}
	constant = \left(1, 2, \frac{2^2}{2!}, \dots, \frac{2^{t-1}}{(t-1)!}\right) \in \mathbb{R}^t 
\end{equation}
\begin{equation}
	s_0 = 0 \in \mathbb{R}^{D \times t} 
\end{equation}
\begin{equation}
	z_0 = 0 \in \mathbb{R}^{D \times t} 
\end{equation}
\begin{equation}
	K_i = \left(1, k_i, k_i^2, \dots, k_i^{t-1}\right) \in \mathbb{R}^{D \times t} 
\end{equation}
\begin{equation}
	Q_i = \left(1, q_i, q_i^2, \dots, q_i^{t-1}\right) \in \mathbb{R}^{D \times t} 
\end{equation}
\begin{equation}
	s_i = s_{i-1} + K_i e^{-k_i^2} v_i 
\end{equation}
\begin{equation}
	z_i = z_{i-1} + K_i e^{-k_i^2} 
\end{equation}
\begin{equation}
	num = sum\left[s_i Q_i constant\right] \in \mathbb{R}^D
\end{equation}
\begin{equation}
	den = sum\left[z_i Q_i constant\right] \in \mathbb{R}^D 
\end{equation}
\begin{equation}
	y_i = \frac{num}{den} 
\end{equation}
where squaring, multiplication and division operations are performed element-wise; $K_i$, $Q_i \in \mathbb{R}^{D\times t}$ and $constant\in\mathbb{R}^t$ can be efficiently generated using the broadcast mechanism; vectors $e^{-{k_i}^2}v_i$, $e^{-{k_i}^2}\in\mathbb{R}^D$ and $constant\in\mathbb{R}^t$ can be element-wise multiplied with high-dimensional matrices via the broadcast mechanism. The caches $s_i$, $z_i\in\mathbb{R}^{D\times t}$ are updated at each step while maintaining constant dimensions, ensuring that the inference cost remains independent of sequence length.

\subsection{Relation to and Differences from Previous Methods}\label{sec:Relation to and Differences from Previous Methods}

Table \ref{complexity comparison} compares the EA-series with typical attention mechanisms from various perspectives, including computational and memory complexity during training, and inference complexity. Generally, the sequence length $L$ is much greater than the feature dimension $D$, and $D$ is significantly larger than the highest order $t$ in the Taylor polynomial. The EA-series demonstrates reduced memory complexity and the lowest computational and inference complexity. Detailed comparisons with specific methods are provided below:

\textbf{SA} In SA, each token vector of dimension $D$ is divided into $H$ heads, where each head is a sub-vector containing $D/H$ channels. The output at each head is computed as follows, with the scaling factor omitted for simplicity: 
\begin{equation}
y_i = \frac{\sum_{j=1}^L e^{q_i k_j^T} v_j}{\sum_{j=1}^L e^{q_i k_j^T}}
\end{equation}
Four key comparisons can be made. (1) The SA uses dot-product operations to compute the similarity, while the EA uses element-wise squared Euclidean distance. (2) The SA generates a total of $H$ head-level feature maps, while the EA generates $D$ channel-level feature maps, allowing for more comprehensive interactions between tokens. (3) The exponential function introduces “spikiness” which contributes much to attention’s effectiveness, while the exponential function incurs significant complexity. We use a Taylor polynomial to approximate the exponential function, preserving its effectiveness while reducing complexity. (4) During inference, SA caches all previous keys and values, causing its inference cost to scale with sequence length. Conversely, the EA-series achieves constant inference complexity with respect to the sequence length, enabling efficient long-sequence inference.
\begin{table}[t]
	\caption{EA-series compares with typical attention mechanisms in terms of computational and memory complexity during training, and inference complexity.}
	\label{complexity comparison} 
	\vskip 0.15in
	\begin{center}
		\begin{small}
			\begin{sc}
				\begin{tabular}{lccc}
					\toprule
					 & Computational & Memory & Inference \\
					\midrule
					SA & $\mathcal{O}(L^2D)$ & $\mathcal{O}(L^2)$ & $\mathcal{O}(LD)$ \\
					LA & $\mathcal{O}(LD^2)$ & $\mathcal{O}(LD)$ & $\mathcal{O}(D^2)$\\
					AFT    & $\mathcal{O}(L^2D)$& $\mathcal{O}(LD)$ & $\mathcal{O}(LD)$\\
					\midrule
					EA-series & $\mathcal{O}(tLD)$ & $\mathcal{O}(tLD)$ & $\mathcal{O}(tD)$ \\
					\bottomrule
				\end{tabular}
			\end{sc}
		\end{small}
	\end{center}
	\vskip -0.1in
\end{table}

\textbf{LA} As shown in equation \ref{equation of LA}, Linear Attention (LA) approximates the $exp(q_ik_j^T)$ in SA with kernels $\varphi(q_i) \cdot \varphi(k_j^T)$, thereby achieving linear training complexity and constant inference complexity with respect to the sequence length. 
\begin{equation}
y_i = \frac{\sum_{j=1}^L \varphi(q_i) \varphi(k_j^T) v_j}{\sum_{j=1}^L \varphi(q_i) \varphi(k_j^T)}
= \frac{\varphi(q_i) \sum_{j=1}^L \varphi(k_j^T) v_j}{\varphi(q_i) \sum_{j=1}^L \varphi(k_j^T)}
\label{equation of LA}
\end{equation}
Two key comparisons can be made. (1) The LA loses the "spikiness" when approximating the exponential function with kernels, leading to performance degradation. In contrast, the EA-series avoids such limitations. (2) Due to the large feature dimension $D$ in LLMs, the $\mathcal{O}(D^2)$ inference complexity of LA results in substantial inference cost. In contrast, the EA-series achieves significantly lower inference cost of an $\mathcal{O}(tD)$ complexity.

\textbf{AFT} AFT \cite{AFT} defines an attention mechanism as follows, where multiplication and division operations are performed element-wise; $w \in \mathbb{R}^{L \times L}$ is learned positional biases. 
\begin{equation}
	y_i = \frac{\sum_{j=1}^L e^{k_j + w_{ij}} v_j}{\sum_{j=1}^L e^{k_j + w_{ij}}}
	\label{equation of AFT}
\end{equation}
Both EA and AFT use element-wise operations. Comparing equation \ref{equation of AFT} with equation \ref{the complete computational formula for EA}, the key difference is that AFT computes attention weights using the position-bias-corrected $k$, while EA computes attention weights based on the similarity between $q$ and $k$.

\textbf{RNNs} Four key comparisons can be made. (1) Traditional RNNs cannot be trained in parallel, resulting in low training efficiency. Linear RNNs enable parallel training by removing nonlinear functions in traditional RNNs, while the removal compromises their representational capacity. In contrast, the EA-series achieves high training efficiency while preserving representational capacity by approximating the nonlinear function with the Taylor polynomial. (2) Unlike direct token interactions in attention mechanisms, RNNs process tokens sequentially, compressing historical information into hidden states. This sequential processing in RNNs gradually diminishes prior information, limiting the model's ability to capture long-range dependencies. (3) Although the autoregressive abilities of RNNs are comparable to those of causal attention mechanisms, RNNs lack the global interaction abilities in non-causal attention mechanisms. (4) Although the inference cost of attention mechanisms is generally significantly higher than those of RNNs, EA-series can be reformulated as recurrent neural networks during inference, achieving an $\mathcal{O}(tD)$ inference cost, which approaches the $\mathcal{O}(D)$ inference cost of RNNs.

\textbf{Hedgehog} Hedgehog \cite{hedgehog} approximates the exponential function in SA using a Taylor polynomial as follows:
\begin{equation}
	e^{q_i k_j^T} = 1 + q_i k_j^T + \frac{(q_i k_j^T)^2}{2} + \cdots
\end{equation}
This approximation reduces the complexity from quadratic to linear with respect to the sequence length. And the experimental results show that the approximation preserves key properties of the exponential function, such as "spikiness" and monotonicity.  

Due to the dot-product operations in SA, approximating $exp(q_ik_j^T)$ introduces exponential complexity with respect to the feature dimension $D$. Specifically, using a $t$-order Taylor polynomial to approximate $exp(q_ik_j^T)$ results in a training complexity of $\mathcal{O}(LD^t)$. This implies that employing high-order Taylor polynomials significantly increases complexity. In contrast, the EA-series, which relies on element-wise operations, achieves a training complexity of $\mathcal{O}(tLD)$, allowing to incorporate sufficient Taylor terms to ensure the effectiveness while maintaining low complexity.

\section{Experiment}

Section \ref{sec:Performance Comparisons} evaluates the performance of EA-2, EA-6, and SA in both causal and non-causal forms, where "2" and "6" indicate the highest order in the Taylor polynomial. Section \ref{sec:Training Cost} analyzes their training efficiency based on memory usage, the BS-L curves, and throughput. Section \ref{sec:Inference Cost} compares their inference cost, focusing on memory usage and latency.

\subsection{Performance Comparisons} \label{sec:Performance Comparisons}

We compare the performance of EA-2, EA-6, and SA in both non-causal and causal forms. The models are constructed using standard Transformer blocks, where Post-Layer Normalization (Post-LN) \cite{post-ln} is used; absolute positional embedding \cite{bert}, which is designed based on the sampling times of the data, is used to incorporate temporal information. To enable direct comparisons between EA-2, EA-6, and SA, we modify only the attention mechanisms within the models while keeping all other components unchanged. The implementation follows the Time Series Library repository \cite{TimesNet}, adhering to its data processing procedures and the specified training, validation, and test splits. To ensure fairness, all models are trained under identical conditions, and the optimal model is selected based on performance on the validation set. Full implementation details are provided on GitHub. 

\textbf{Non-causal form} In non-causal attention mechanisms, each token can attend to all tokens in the sequence. Consequently, non-causal attention is well-suited for tasks involving understanding, summarization, and classification. We compare the performance of the non-causal EA-2, EA-6 and SA on multivariate time series classification (MTSC). In MTSC, each sample consists of multiple interrelated time series, with each series containing observations recorded at sampling times. The objective is to predict the label of each sample. Specifically, a sample $S=\left(S_1,\ldots,S_n\right)\in\mathbb{R}^{L\times n}$ comprises $n$ time series, each of length $L$. The model's goal is to predict the label associated with $S$. We choose four real-world MTSC datasets from the UEA Time Series Classification Archive \cite{UEA}: JapaneseVowels, SelfRegulationSCP1, SelfRegulationSCP2 and UWaveGesture. Table \ref{the characteristics of MTSC} provides the characteristics of these datasets, including the number of time series, the length of each time series and the number of labels. Table \ref{classification results} reports the models’ prediction accuracy on these datasets, where higher accuracy indicates better performance. The results show that the EA-series underperforms with few Taylor terms. With a sufficient number of terms, the EA-series demonstrates strong performance, surpassing that of SA.

\begin{table}[t]
	\caption{The characteristics of multivariate time series classification datasets, including the number of time series, the length of each time series and the number of labels.}
	\label{the characteristics of MTSC} 
	\vskip 0.15in
	\begin{center}
		\begin{small}
			\begin{sc}
				\begin{tabular}{lcccc}
					\toprule
					& JAP & SCP1 & SCP2 & UWG \\
					\midrule
					\# of series & 12 & 6 & 7 & 3 \\
					Length of series & 29 & 896 & 1152 & 315 \\
					\# of labels      & 9  & 2 & 2 & 8 \\
					\bottomrule
				\end{tabular}
			\end{sc}
		\end{small}
	\end{center}
	\vskip -0.1in
\end{table}

\begin{table}[t]
	\caption{Multivariate time series classification results. Bolded values indicate the best results.}
	\label{classification results} 
	\vskip 0.15in
	\begin{center}
		\begin{small}
			\begin{sc}
				\begin{tabular}{lcccc}
					\toprule
					& JAP & SCP1 & SCP2 & UWG \\
					\midrule
					EA-2 & 0.957 & 0.887 & 0.500 & 0.794 \\
					EA-6 & \textbf{0.973} & \textbf{0.904} & \textbf{0.533} & \textbf{0.847} \\
					SA   & 0.970 & 0.894 & 0.528 & 0.822 \\
					\bottomrule
				\end{tabular}
			\end{sc}
		\end{small}
	\end{center}
	\vskip -0.1in
\end{table}

\begin{table*}[!t]
	\caption{Time series forecasting results. Bolded values indicate the best results}
	\label{forecasting results} 
	\vskip 0.15in
	\begin{center}
		\begin{small}
			\begin{sc}
				\begin{tabular}{lcccccccccccc}
					\toprule
					\multicolumn{1}{c}{} & \multicolumn{4}{c}{ETTH2} & \multicolumn{4}{c}{ETTM2} & \multicolumn{4}{c}{TRAFFIC} \\
					\cmidrule(lr){2-5} \cmidrule(lr){6-9} \cmidrule(lr){10-13}
					& \multicolumn{2}{c}{6} & \multicolumn{2}{c}{12} & \multicolumn{2}{c}{6} & \multicolumn{2}{c}{12} & \multicolumn{2}{c}{6} & \multicolumn{2}{c}{12} \\
					\cmidrule(lr){2-3} \cmidrule(lr){4-5} \cmidrule(lr){6-7} \cmidrule(lr){8-9} \cmidrule(lr){10-11} \cmidrule(lr){12-13}
					& MAE & RMSE & MAE & RMSE & MAE & RMSE & MAE & RMSE & MAE & RMSE & MAE & RMSE \\
					\midrule
					EA-2 & 1.43 & 2.01 & 2.14 & 2.90 & 0.44 & 0.63 & 0.73 & 1.08 & 0.12 & 0.15 & 0.18 & 0.28 \\
					EA-6 & \textbf{1.23} & \textbf{1.79} & \textbf{1.90} & \textbf{2.67} & \textbf{0.36} & \textbf{0.53} & \textbf{0.64} & \textbf{0.98} & \textbf{0.09} & \textbf{0.13} & \textbf{0.16} & \textbf{0.25} \\
					SA   & 1.24 & \textbf{1.79} & 1.92 & 2.71 & \textbf{0.36} & 0.54 & \textbf{0.64} & 0.99 & 0.10 & \textbf{0.13} & \textbf{0.16} & 0.27 \\
					\bottomrule
				\end{tabular}
			\end{sc}
		\end{small}
	\end{center}
	\vskip -0.1in
\end{table*}

\begin{figure*}[!t]
	\captionsetup[subfigure]{labelformat=simple}
	\renewcommand\thesubfigure{(\alph{subfigure})}
	\centering
	\begin{subfigure}{0.33\textwidth}%
		\centering
		\includegraphics[width=\linewidth]{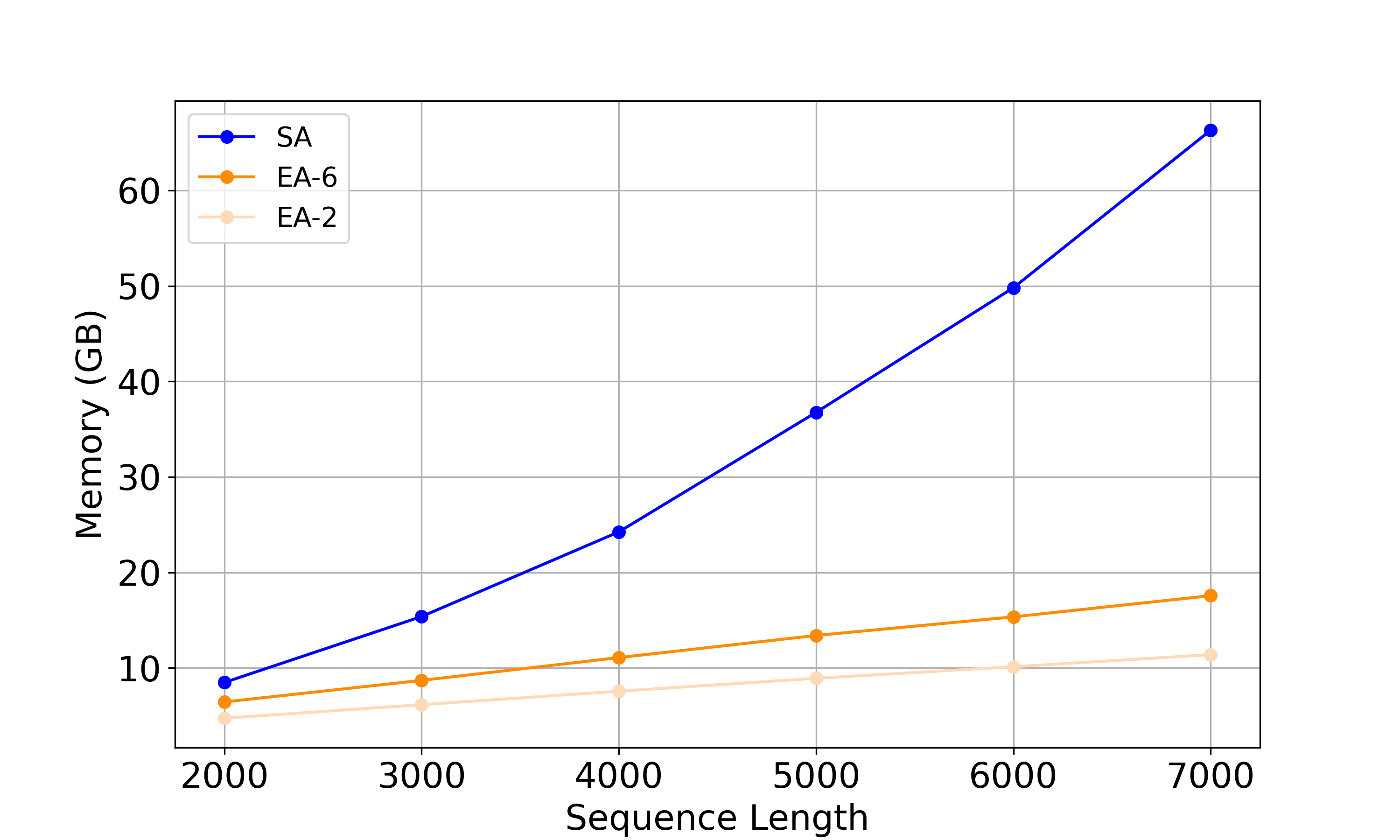}
		\caption{}
		\label{TrainCost_L-Memory}%文中引用该图片代号
	\end{subfigure}
	\centering
	\begin{subfigure}{0.33\textwidth}
		\centering
		\includegraphics[width=\linewidth]{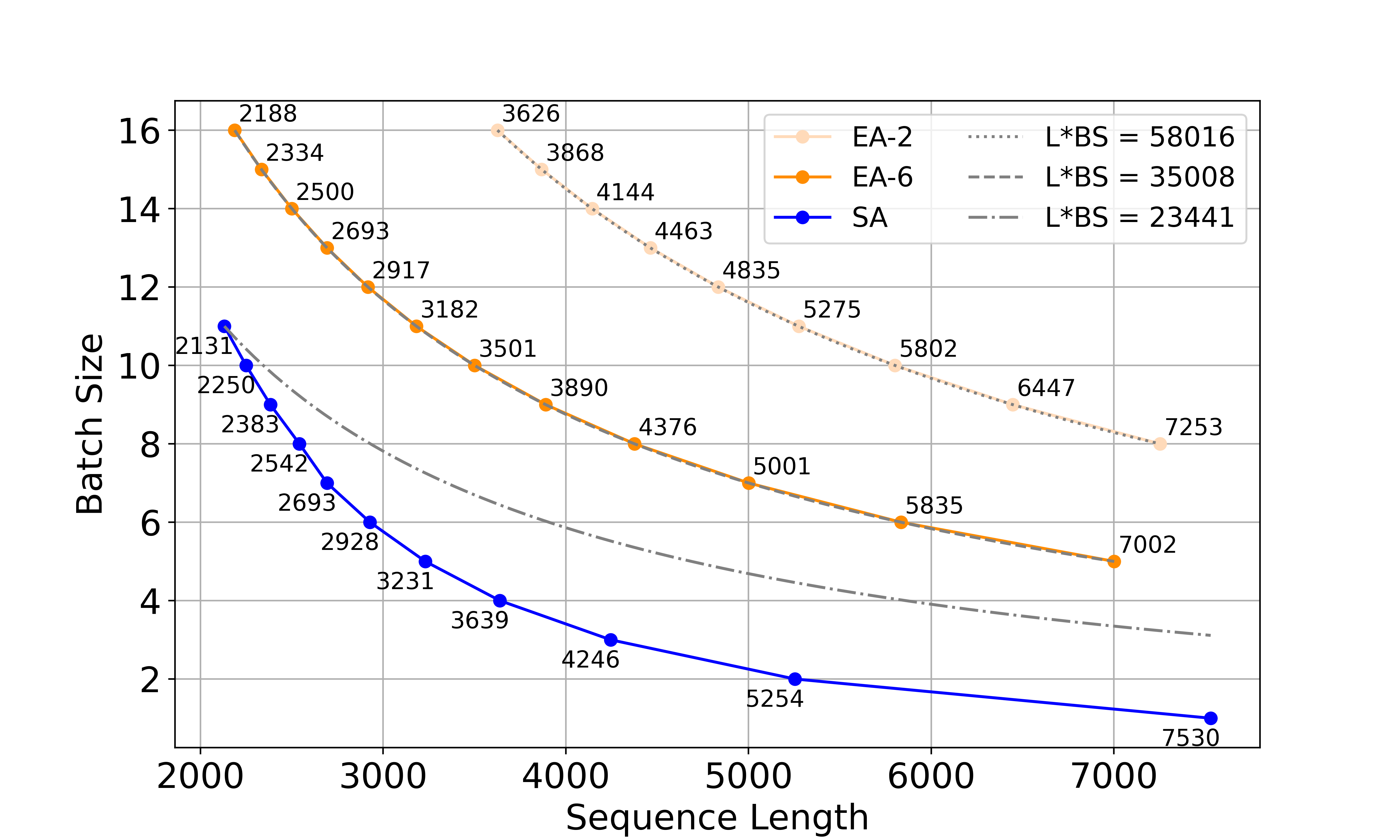}
		\caption{}
		\label{TrainCost_Length-BS}%文中引用该图片代号
	\end{subfigure}
	\centering
	\begin{subfigure}{0.33\textwidth}
		\centering
		\includegraphics[width=\linewidth]{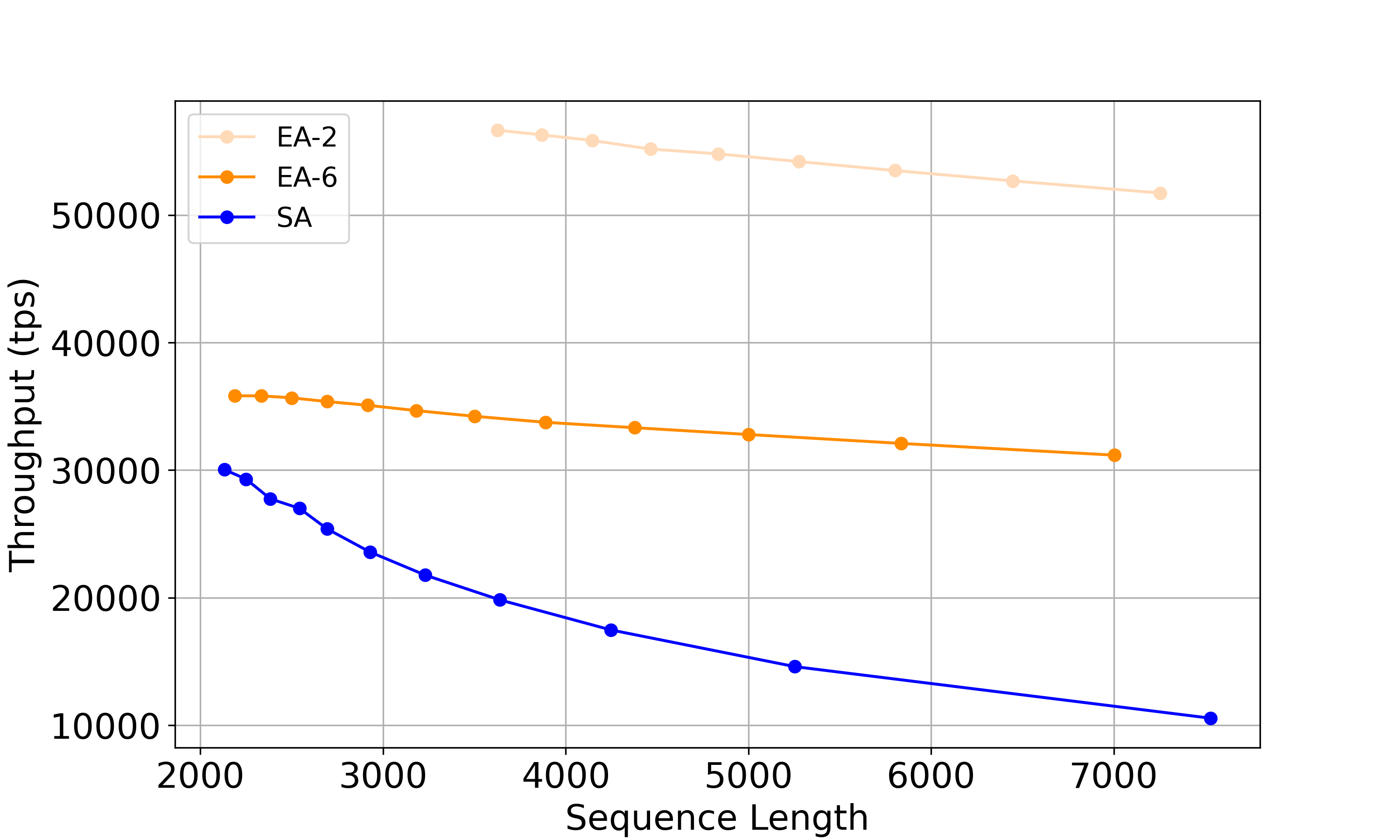}
		\caption{}
		\label{TrainCost_L-Throughput}%文中引用该图片代号
	\end{subfigure}
	\caption{Training cost of EA-2, EA-6, and SA. Specifically, (a) illustrates their memory usage, (b) presents their BS-L curves, and (c) shows their throughput.}
	\label{TrainCost}
\end{figure*}

\textbf{Causal form} In causal attention mechanisms, each token attends only to preceding and current tokens. Consequently, causal attention is well-suited for tasks involving generation and forecasting. We compare the performance of the causal EA-2, EA-6 and SA on time series forecasting (TSF). In TSF, the model's objective is to predict the future series based on a given series. Specifically, for a time series $S=\left(s_1,\ldots,s_L\right)\in\mathbb{R}^L$, the model aims to forecast the future series $S^\prime=\left(s_{L+1},\ldots,s_{L+L^\prime}\right)\in\mathbb{R}^{L^\prime}$. We evaluate the models using three real-world TSF datasets: ETTh2, ETTm2, and Traffic. In the experiments, we set $L=6$ and assess the models’ performance of $L^\prime=6$ and $L^\prime=12$ using the Mean Absolute Error (MAE) and Root Mean Squared Error (RMSE) metrics, where lower metric values indicate better performance. Table \ref{forecasting results} presents the time series forecasting results. The results shows that the EA-series underperforms with few Taylor terms. As more terms are incorporated, the EA series exhibits superior performance, outperforming the SA.

\subsection{Training Cost} \label{sec:Training Cost}

In this section, we compare the training cost of EA-2, EA-6, and SA. All models use the BERT-base \cite{bert} configuration with 12 layers, a hidden dimension $D=768$, a head dimension of 64, and an intermediate dimension of $4D$ in the FFN. The models are implemented in PyTorch without additional optimizations, and experiments are conducted on an A800-80GB GPU.

\textbf{Memory} Figure \ref{TrainCost_L-Memory} illustrates the memory usage of the models at different sequence lengths with a batch size of 1. The memory usage of the EA-series increases linearly with the sequence length, due to its linear memory complexity with respect to the sequence length. In contrast, the memory usage of SA grows exponentially due to its quadratic memory complexity. Overall, the EA-series requires substantially less memory than SA for long-sequence modeling.

\textbf{BS-L Curve} Maximizing GPU memory utilization is important for model training. For a given model and GPU, memory utilization is primarily influenced by the batch size (BS) and sequence length (L). In figure \ref{TrainCost_Length-BS}, the solid curves, referred to as the BS-L curves, depict the maximum sequence lengths the models can handle on the GPU for various BS. Clearly, BS and L have an inverse relationship: as the sequence length increases, the batch size that fits on the GPU decreases. The dashed curves are inverse proportional curves. The product of BS and L, representing the number of tokens processed per step, remains constant along each dashed curve.

Key observations from the figure are as follows. (1) The EA-series can process more tokens per step than SA due to the lower memory usage of EA-series, making EA-series advantageous for processing large datasets. (2) The BS-L curves of the EA-series closely align with the dashed curves, whereas that of SA deviates downward from the dashed curve as the sequence length increases. This implies that the total number of tokens SA can process decreases as sequence length increases, leading to reduced efficiency in long-sequence modeling.

\begin{figure*}[ht]
	\captionsetup[subfigure]{labelformat=simple}
	\renewcommand\thesubfigure{(\alph{subfigure})}
	\centering
	\begin{subfigure}{\linewidth}
		\centering
		\includegraphics[width=\linewidth]{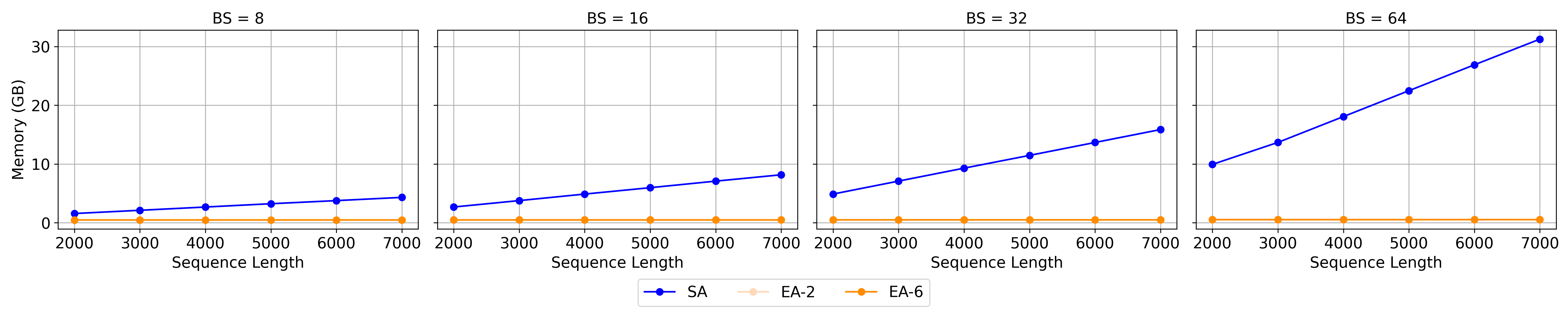}
		\caption{}
		\label{InferenceCost_L-Memory}%文中引用该图片代号
	\end{subfigure}
	\centering
	\begin{subfigure}{\linewidth}
		\centering
		\includegraphics[width=\linewidth]{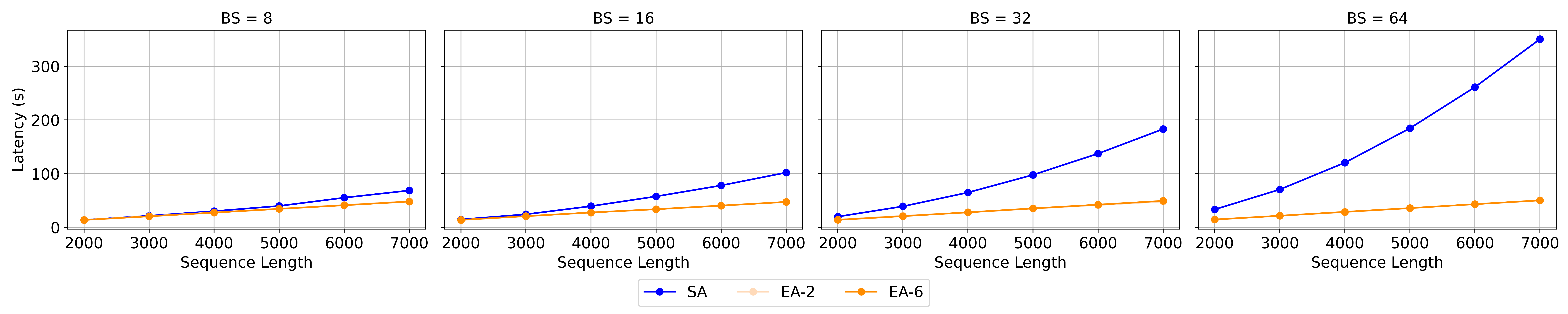}
		\caption{}
		\label{InferenceCost_L-Latency}%文中引用该图片代号
	\end{subfigure}
	\centering
	\caption{Inference cost of EA-2, EA-6, and SA. Specifically, (a) illustrates their memory usage, and (b) presents their latency.}
	\label{inference cost}
\end{figure*}

\textbf{Throughput} Throughput refers to the amount of data a model can process within a given time frame, typically measured in tokens per second (tps). Figure \ref{TrainCost_L-Throughput} presents the throughput at various points along the BS-L curves. Key observations from the figure are as follows. (1) The EA-series achieves significantly higher throughput than SA. (2) The EA-series maintains high throughput, while SA's throughput decreases notably with longer sequences.

\subsection{Inference Cost} \label{sec:Inference Cost}

In this section, we compare the inference cost of EA-2, EA-6, and SA in terms of memory usage and latency. All models use the BERT-base configuration, and all experiments are conducted on an A800-80GB GPU. During inference, SA utilizes KV-caching to improve efficiency, resulting in an $\mathcal{O}(LD)$ inference complexity; EA-series employs the recurrent representation described in section \ref{sec:Causal EA-series}, achieving an $\mathcal{O}(tD)$ inference complexity. Figure \ref{inference cost} illustrates the performance of the models. From the figure, we observe that EA-2 and EA-6 exhibit very similar performance, despite EA-6 containing more terms than EA-2; EA-series achieves significantly better inference efficiency than SA. Detailed comparisons are presented below. 

\textbf{Memory} Figure \ref{InferenceCost_L-Memory} illustrates the memory usage of the models under varying batch sizes and sequence lengths. The memory usage of SA increases linearly with both sequence length and batch size. In contrast, the memory usage of EA-series remains constant with respect to the sequence length, because the size of the caches $s_i,z_i\in\mathbb{R}^{D\times t}$ is independent of the sequence length; the memory usage of EA-series is minimally affected by changes in batch size, because the size of the caches remains negligible compared to the model parameters, even for large batch sizes. 

\textbf{Latency} Figure \ref{InferenceCost_L-Latency} illustrates the latency of the models under varying batch sizes and sequence lengths. Key observations from the figure are as follows. (1) The latency of the EA-series increases linearly with the number of tokens, while the latency of the SA increases exponentially. This occurs because the cache size of the EA-series remains constant with respect to the sequence length, resulting in a fixed generation time per token. In contrast, the cache size of the SA grows with the sequence length, leading to progressively longer generation time for subsequent tokens. (2) The latency of the EA-series is minimally affected by batch size changes, whereas the latency of the SA is significantly influenced. This is because the EA-series maintains a small cache size even with large batch sizes, resulting in minimal variation in computation time. In contrast, the SA has a large cache size, and increasing the batch size significantly increases computation time.

\section{Conclusion}

In this work, we propose an element-wise attention mechanism as a replacement for SA. The mechanism substitutes the dot-product operation with element-wise operation for similarity computation and approximates the quadratic complexity term using a Taylor polynomial. By doing so, it simultaneously achieves SA-comparable performance and incredible efficiency (a training complexity of $\mathcal{O}(tLD)$ and an inference complexity of $\mathcal{O}(tD)$. The proposed model offers a foundation for the next-generation architecture and may have significant impact across various domains.

\newpage

\newpage

\section*{Impact Statement}

This paper presents work whose goal is to advance the field of 
Machine Learning. There are many potential societal consequences 
of our work, none which we feel must be specifically highlighted here.


\bibliography{example_paper}
\bibliographystyle{icml2025}

%%%%%%%%%%%%%%%%%%%%%%%%%%%%%%%%%%%%%%%%%%%%%%%%%%%%%%%%%%%%%%%%%%%%%%%%%%%%%%%
%%%%%%%%%%%%%%%%%%%%%%%%%%%%%%%%%%%%%%%%%%%%%%%%%%%%%%%%%%%%%%%%%%%%%%%%%%%%%%%
% APPENDIX
%%%%%%%%%%%%%%%%%%%%%%%%%%%%%%%%%%%%%%%%%%%%%%%%%%%%%%%%%%%%%%%%%%%%%%%%%%%%%%%
%%%%%%%%%%%%%%%%%%%%%%%%%%%%%%%%%%%%%%%%%%%%%%%%%%%%%%%%%%%%%%%%%%%%%%%%%%%%%%%

\end{document}